\documentclass{article}

     \usepackage[preprint,nonatbib]{neurips_2022}

\usepackage[utf8]{inputenc} 
\usepackage[T1]{fontenc}    
\usepackage{url}            
\usepackage{booktabs}       
\usepackage{amsfonts}       
\usepackage{nicefrac}       
\usepackage{xcolor}         
\usepackage{bm}
\usepackage{amsmath}
\usepackage[frozencache]{minted}
\usepackage{caption}
\usepackage{subcaption}
\usepackage{graphicx}
\usepackage{svg}
\usepackage{tabularx}
\usepackage{adjustbox}
\usepackage{array, booktabs, makecell}
\usepackage{tablefootnote}
\usepackage{siunitx}
\usepackage[title]{appendix}
\usepackage{longtable}
\usepackage{multirow}

\usepackage[
activate   = {true},
protrusion = false,
expansion  = true,
kerning    = true,
spacing    = true,
tracking   = false,
auto       = true,
selected   = true,
factor     = 1000,
stretch    = 15,
shrink     = 15,
]{microtype}

\usepackage[
maxbibnames=99,
maxcitenames=2,
natbib=true,
style=numeric-comp,
backend=biber,
sorting=none,
giveninits=true,
url=false,
doi=false,
isbn=false,
]{biblatex}
\definecolor{gk}{RGB}{120, 120, 120}
\definecolor{gg}{HTML}{5f9411}
\definecolor{gb}{HTML}{417598}
\definecolor{gr}{HTML}{d15120}
\definecolor{gy}{HTML}{d2ad00}
\definecolor{LightGray}{gray}{0.925}

\usepackage[pdfa,colorlinks,bookmarksopen,bookmarksnumbered,allcolors=gg]{hyperref}

\usepackage[nameinlink,capitalise]{cleveref}
\crefname{line}{line}{lines}
\crefname{figure}{Fig.}{Figs.}
\Crefname{figure}{Fig.}{Figs.}
\crefname{equation}{Eq.}{Eqs.}
\Crefname{equation}{Eq.}{Eqs.}
\crefname{section}{Sec.}{Secs.}
\Crefname{section}{Sec.}{Secs.}
\crefname{definition}{Def.}{Defs.}
\Crefname{definition}{Def.}{Defs.}
\crefname{algorithm}{Alg.}{Algs.}
\Crefname{algorithm}{Alg.}{Algs.}

\AtBeginEnvironment{appendices}{\crefalias{section}{appendix}}
\AtBeginEnvironment{appendices}{\crefalias{subsection}{appendix}}

\newcommand{\code}[1]{{\setlength{\fboxsep}{0pt}\colorbox{LightGray}{\texttt{#1}}}}
\newcommand{\codeg}[1]{{\setlength{\fboxsep}{0pt}\colorbox{LightGray}{\texttt{#1}}}}
\newcommand{\codes}[1]{{\setlength{\fboxsep}{0pt}{\footnotesize \colorbox{LightGray}{\texttt{#1}}}}}
\newcommand{\codess}[1]{{\setlength{\fboxsep}{0pt}{\scriptsize \colorbox{LightGray}{\texttt{#1}}}}}

\newcolumntype{Y}{>{\centering\arraybackslash}X}

\newcommand\blfootnote[1]{%
  \begingroup
  \renewcommand\thefootnote{}\footnote{#1}%
  \addtocounter{footnote}{-1}%
  \endgroup
}

\AtBeginEnvironment{appendices}{\crefalias{section}{appendix}}

\title{The Planner Optimization Problem:\\Formulations and Frameworks}

\author{
  Yiyuan Lee\(^1\) \quad
  Katie Lee\(^1\) \quad
  Panpan Cai\(^2\) \quad
  David Hsu\(^3\) \quad
  Lydia E. Kavraki\(^1\)\\\\
  \(^1\)Rice University \quad
  \(^2\)Shanghai Jiao Tong University \quad
  \(^3\)National University of Singapore \\\\
  {
    \footnotesize
  \texttt{\{yiyuan.lee, katie.lee, kavraki\}@rice.edu}
  }
}

\AtBeginEnvironment{appendices}{\crefalias{section}{appendix}}

\begin{document}

\maketitle

\begin{abstract}
  \small
  Identifying internal parameters for planning is crucial to maximizing the performance of a planner. However, automatically tuning internal parameters which are conditioned on the problem instance is especially challenging. A recent line of work focuses on learning planning parameter \emph{generators}, but lack a consistent problem definition and software framework. This work proposes the unified \emph{planner optimization problem} (POP) formulation, along with the \emph{Open Planner Optimization Framework} (OPOF), a highly extensible software framework to specify and to solve these problems in a reusable manner.
\end{abstract}

\section{Introduction}

Planning \emph{is} a cornerstone of decision making in artificial intelligence. In planning, a planner is given an instance of a problem and utilizes some form of computation to realize a certain goal. Often, the performance of the planner is heavily influenced by a set of internal \emph{planning parameters}. In order to maximize the performance of the planner, it is crucial to select the best parameters. 
In this work, we assume the general case where high-quality parameters are hard to derive analytically, especially for highly complex planners. Instead, we focus on how to \emph{tune} the parameters using a selected set of training problem instances. This essentially treats the planner as a black-box function, and we try out different parameters until we are satisfied with the observed performance of the planner. Methods developed for such a paradigm are extremely general, since they make no assumption about the internals of the planner.

To \emph{automatically} tune a given planner, prevailing methods in what is known as algorithm configuration (AC) \cite{hutter2009paramils, lopez2016irace, ansotegui2009gender, lopez2016irace} or hyperparameter optimization (HPO) \cite{feurer2019hyperparameter} typically find a fixed set of parameters which work best on average across the training set of problem instances. However, we argue that the parameters should be \emph{conditioned} on the problem instance to achieve maximal performance. In particular, we need to identify a \emph{generator} which maps a given problem instance to high-quality planning parameters. This is extremely challenging because the space of such possible mappings is much larger and even less understood than the space of planning parameters. An emerging trend of work focuses on some form of this problem -- for example by learning belief-dependent macro-actions for online POMDP planners \cite{lee2021a}; learning workspace- and task-dependent sampling distributions for sampling-based motion planners (SBMPs) \cite{lee2022a}; and learning belief-dependent attention for autonomous driving \cite{danesh2022a}. While the key ideas and techniques behind these works are almost exactly the same, they lack a consistent problem formulation and have vastly different codebases.

In this work, we introduce the general \emph{planner optimization problem} (POP) formulation which unifies the key ideas behind such recent contributions in tuning planning parameters that are conditioned on the problem instance. Additionally, we provide a highly extensible software framework, the \emph{Open Planner Optimization Framework} (OPOF), to specify and to solve these problems in a reusable manner. We hope that this will allow the community to approach the problem in a coherent and accelerated manner. OPOF is available open-source at \href{https://opof.kavrakilab.org}{https://opof.kavrakilab.org}.

\section{Background}

\subsection{Planner Optimization Problem (POP)}

A \emph{planner optimization problem} (POP) is a \(4\)-tuple \((\mathcal{C}, \mathcal{X}, \boldsymbol{f}, \mathcal{D})\), where \(\mathcal{C}\) is the \emph{problem instance space}, \(\mathcal{X}\) is the \emph{planning parameter space}, \(\boldsymbol{f}\) is the \emph{planning objective}, and \(\mathcal{D}\) is the \emph{problem instance distribution}. A \emph{planner} takes as input a problem instance \(c \in \mathcal{C}\) and planning parameters \(x \in \mathcal{X}\), and performs some computation on \(c\) using \(x\). It returns a numeric value indicating the quality of the computation, which inherently follows the distribution \(\boldsymbol{f}(x; c)\), the planning objective. 

The goal is to find a \emph{generator} \(G_{\theta}(c)\), mapping problem instances in \(\mathcal{C}\) to planning parameters in \(\mathcal{X}\), such that the \emph{expected planning performance} is maximized:
\begin{equation}\label{eqn:pop}
   \underset{\theta}{\arg\max} ~ \mathbb{E}_{c \sim \mathcal{D}}[\mathbb{E}_{x \sim G_\theta(c)}[\mathbb{E}[\boldsymbol{f}(x; c)]]].
\end{equation}
The generator may be \emph{stochastic}, in which case samples \(x \sim G_\theta(c)\) are produced whenever the generator is called. On the other hand, a \emph{deterministic} generator always returns the same value \(x = G_\theta(c)\) for the same problem instance \(c\). An \emph{unconditional} generator is one whose output is unaffected by its input \(c\).

The problem is challenging because the generator \(G_\theta(c)\) needs to be learned solely through interaction with the planner. In particular, we only have access to the planner which produces samples of \(\boldsymbol{f}(x; c)\). We do not assume to have an analytical, much less differentiable, form for \(\bm{f}(x; c)\).

\subsection{Related work}

\textbf{Black-box optimization (BBO).} In black-box optimization (BBO) \cite{audet2017a}, we are tasked with finding inputs that maximize a \emph{closed} objective function. In particular, we have no knowledge about the analytical expression or derivatives of the function, and can only evaluate the objective function at different inputs. There are \(2\) broad classes of approaches to solve BBOs. (i) \emph{Evolutionary algorithms} (ES) \cite{grefenstette1986optimization,hansen2003reducing,fortin2012deap,olson2016evaluation,awad2021dehb} track a population of points across the input space. The population is modified based on some notion of mutation, crossover, and culling across iterative evaluations against the objective function. (ii) \emph{Bayesian optimization} (BO) methods \cite{snoek2012practical, hutter2011sequential, bergstra2011algorithms, springenberg2016bayesian, lindauer2022smac3} build a suitable model of the objective function, based on already observed evaluations, and uses the model to acquire subsequent evaluation points in a manner that trades off exploration and exploitation. 

\textbf{Algorithm configuration (AC).} \emph{Algorithm configuration} (AC) \cite{hutter2009paramils, lopez2016irace, ansotegui2009gender, lopez2016irace}, also known as \emph{hyperparameter optimization} (HPO) \cite{feurer2019hyperparameter}, is a form of BBO where we want inputs that work well \emph{on average} over a set of problem instances. Additionally, the importance of generalization is considered -- the inputs optimized for the \emph{training set} must also work well on a \emph{validation set} of unseen problem instances. A typical approach is to treat the averaged objective function as the objective function and to solve the derived BBO problem, for which BBO techniques can be applied \cite{loshchilov2016cma, hinz2018speeding, moll2021hyperplan}.

\textbf{Planner optimization problem (POP).} Our POP formulation (\cref{eqn:pop}) is effectively an extension of AC. In a POP, we seek to find a generator mapping problem instances to high-quality inputs for the objective function. This is in contrast to AC, which seeks only a single input that works well on average across the training set. Such conditional solutions are important in AI planning problems, where the optimal planning parameters depend critically on the problem instance. Different forms of POPs have only been explored recently by works such as \cite{lee2021a,lee2022a,danesh2022a}, which apply the \emph{Generator-Critic} algorithm introduced in \cite{lee2021a}. A deep neural network is used for the generator, which is trained with the help of a \emph{critic}, another deep neural network. The critic is trained to model the response of the planner via supervision loss. It serves as a \emph{differentiable surrogate objective} for the planning objective, enabling training gradients to flow into the generator for gradient-based updates. Through the POP formulation, we hope to unify the key ideas motivating these works.

\textbf{Software Frameworks.} POPs require specialized components to support the notion of a generator and to integrate gradient-based deep learning techniques, such as those used in the aforementioned Generator-Critic algorithm. While software frameworks for BBO such as PyPop7 (ES) \cite{duan2022pypop7} and SMAC3 (BO) \cite{lindauer2022smac3} exist and are actively maintained, rewiring them for POPs would require awkward hacks that would make usage unnatural. Thus, through OPOF, we hope to provide a software framework specially for POPs with intuitive and reusable interfaces, while harnessing the power of deep learning methods exposed via PyTorch \cite{paszke2019pytorch}.


\section{OPOF: Open-Source Planner Optimization Framework}

\subsection{\emph{Domains}: abstracting the planner optimization problem}

At the core of OPOF lies the \emph{domain} abstraction for specifying a planner optimization problem. The \code{Domain} interface consists of a minimal set of functions to specify the details of a given planner optimization problem -- such as the training distribution, the parameter space, and the planner.

\begin{figure}[h]
\begin{minted}[
  bgcolor=LightGray,
  fontsize=\footnotesize
]{python}
domain = RandomWalk2D(11) # Instantiate domain instance.
problem_set = domain.create_problem_set()
planner = domain.create_planner()
problems = [problem_set() for _ in range(100)] # Sample problem instances.
# Sample 100 sets of parameters.
parameters = [ps.rand(100).numpy() for ps in domain.composite_parameter_space()]

for i in range(100):
    result = planner(problems[i], [p[i] for p in parameters], []) # Invoke planner.
    print(result["objective"]) # Print result.
\end{minted}
\caption{Code example demonstrating usage of the \code{Domain} abstraction interface. We show how the key functions of an implemented domain are used. Here, we randomly generate \(100\) sets of problem instances \(c_i\) and planning parameters \(x_i\), and probe the planner for corresponding samples of the planning objective \(\boldsymbol{f}(x_i; c_i)\).}
\label{fig:usage}
\end{figure}

\subsection{Available domains}

\begin{figure}[h]
  \centering
   \includesvg[height=2.5cm]{figures/random_walk2d.svg}
   \includesvg[height=2.5cm]{figures/maze2d.svg}
   \includegraphics[height=2.5cm]{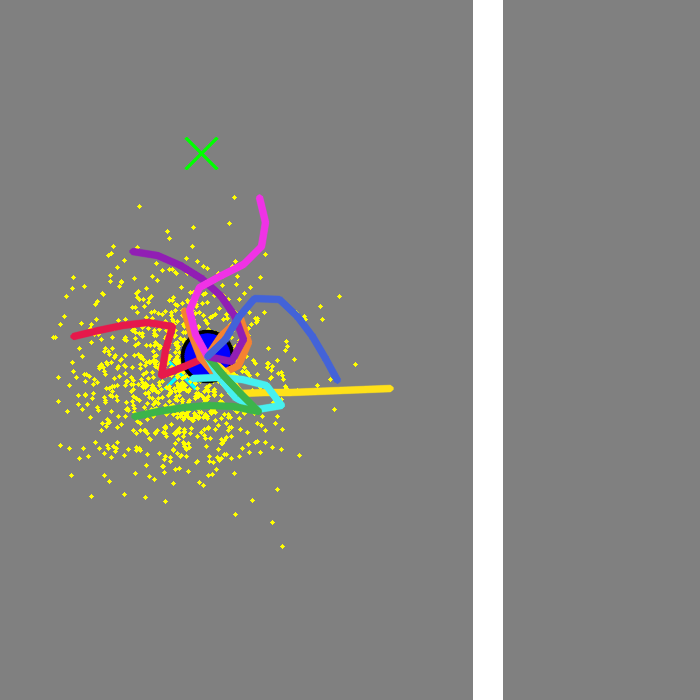}
   \includegraphics[height=2.5cm]{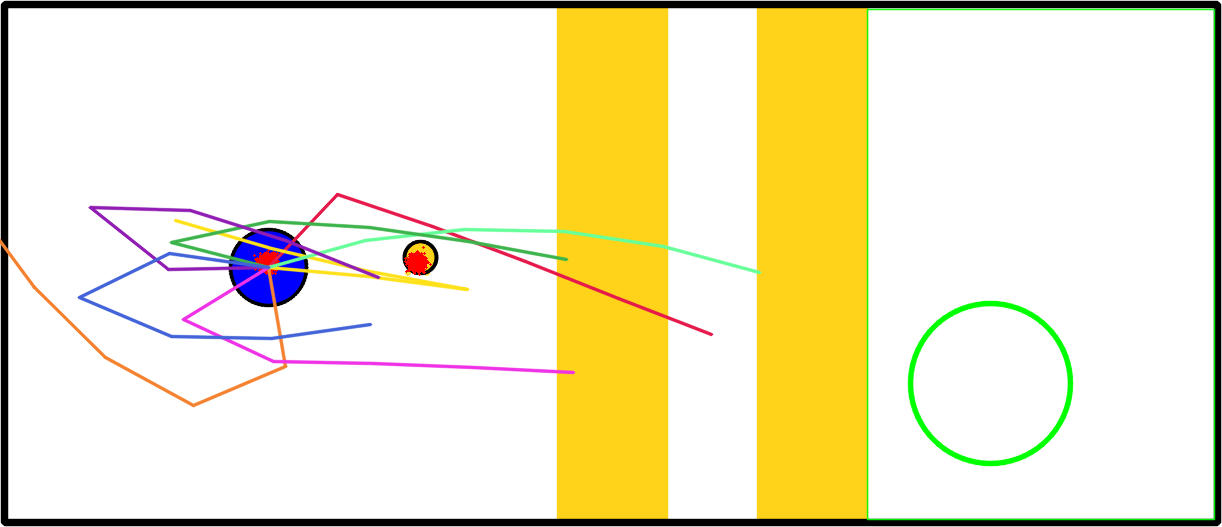}\\
   \includegraphics[height=3cm]{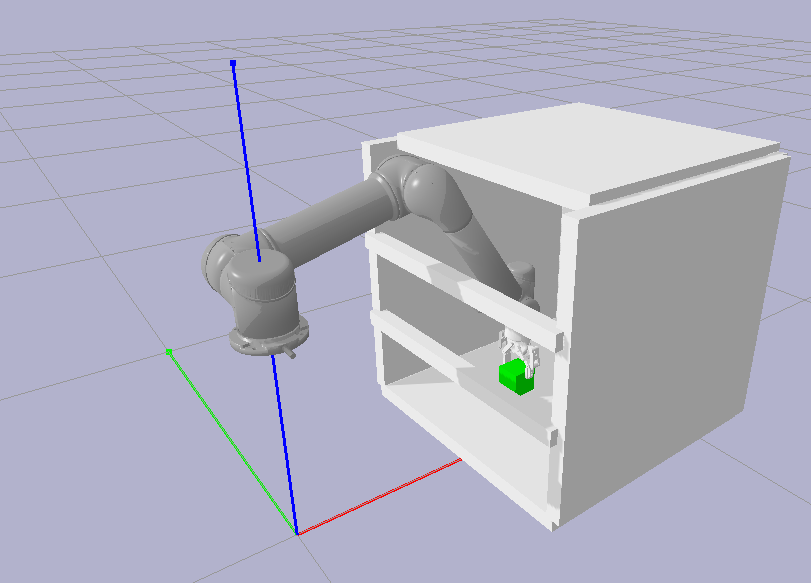}
   \includegraphics[height=3cm]{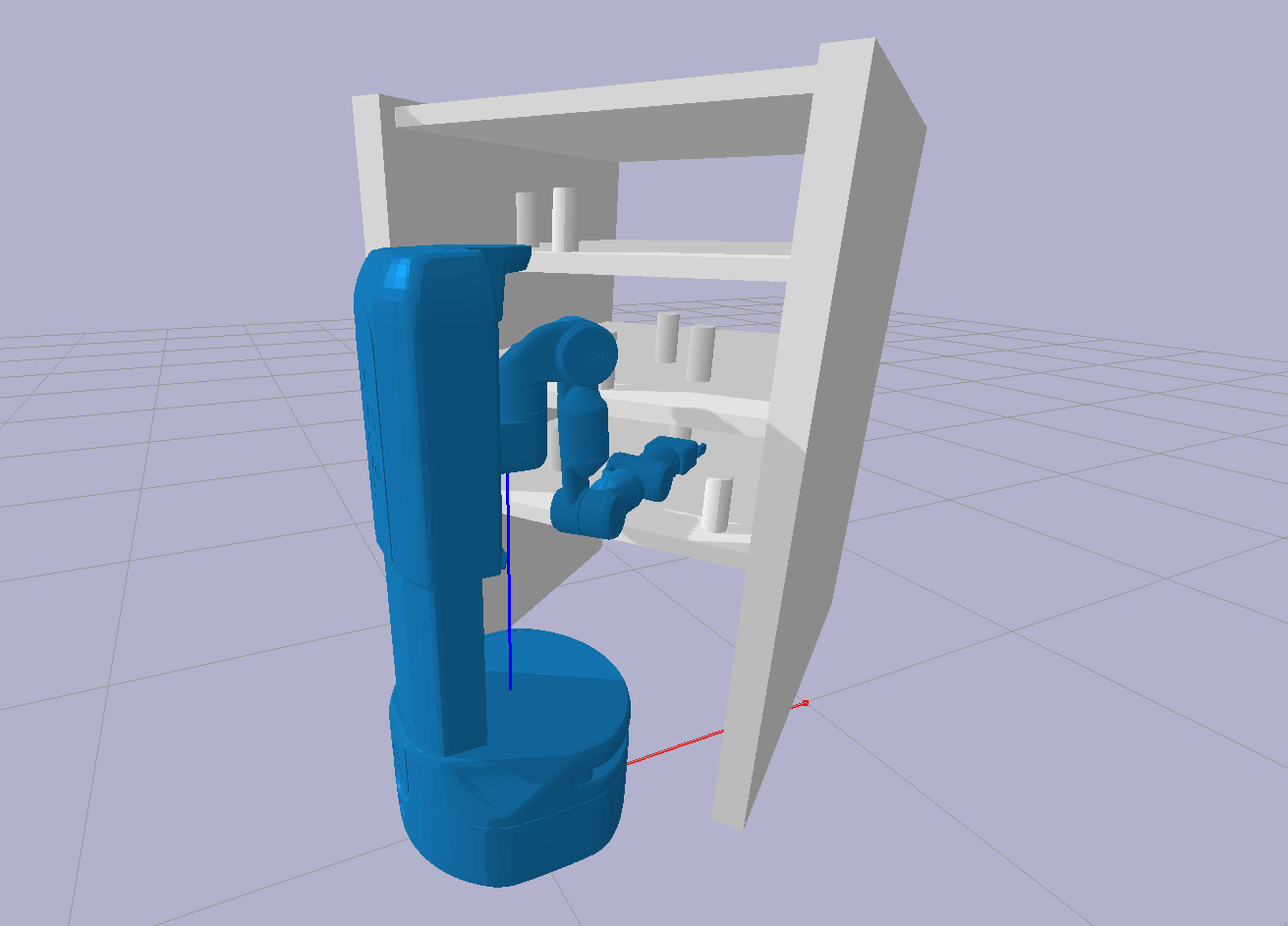}
   \includegraphics[height=3cm]{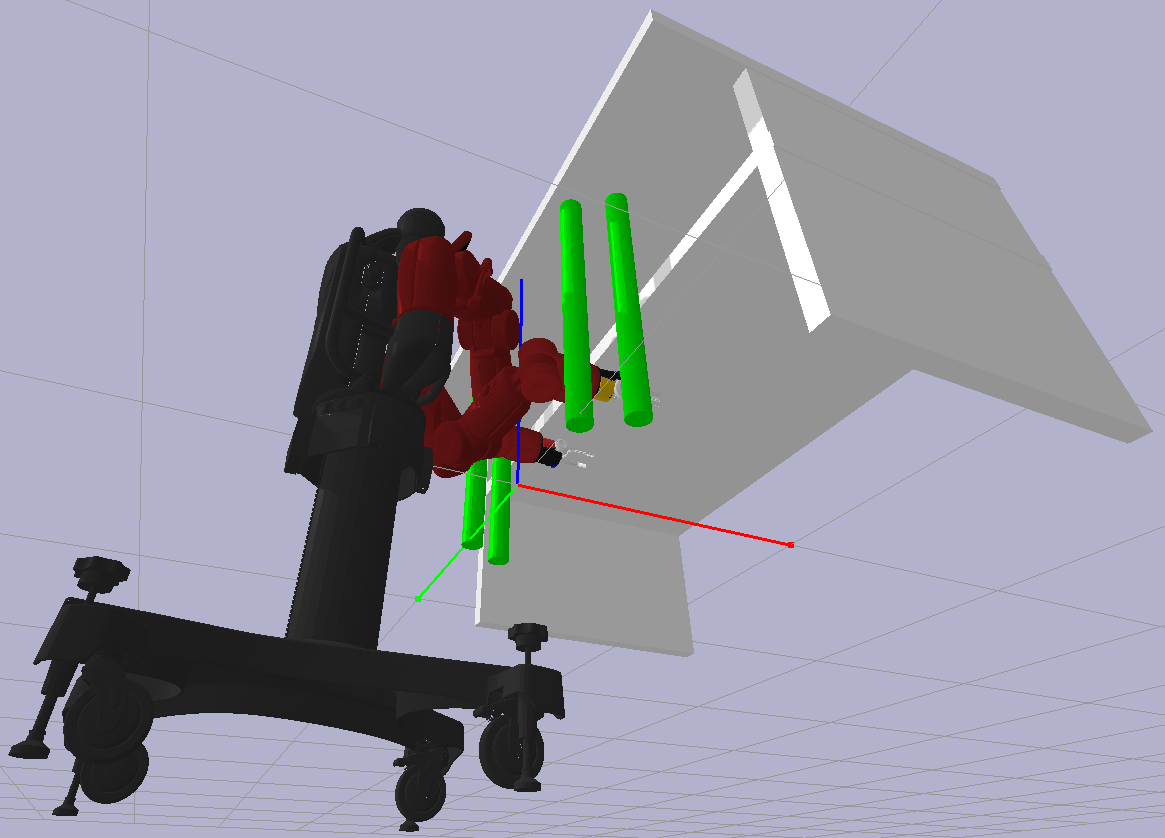}
   \caption{Visualizations of some domains currently available alongside OPOF.}
   \label{fig:domains}
\end{figure}

Domains are implemented in external \code{Pip} packages, which are installed alongside OPOF. We provide a collection of domains (\cref{fig:domains}), which we expect to grow over time. At the time of OPOF's beta release, the following domains are available:

\begin{itemize}
  \item \textbf{2D grid world} (\cref{desc:grid2d})\textbf{.} Small-scale toy domains with easy to understand behavior from classic, discrete planning. Serves as a sanity check against developing algorithms.
  \item \textbf{SBMP} (\cref{desc:sbmp})\textbf{.} Domains from \cite{chamzas2021motionbenchmaker,lee2022a} to optimize hyperparameters, samplers, and projections for sampling-based motion planning. The motion planner is given a tight time budget (\(1\) - \(3\) s), and requires effective parameters to plan reasonably well.
  \item \textbf{Online POMDP planning} (\cref{desc:pomdp})\textbf{.} Domains from \cite{lee2021a} to learn macro-actions for improving online POMDP planning. The POMDP planner is given a very tight time budget (\(100\) ms), and discovering effective macro-actions is critical to plan reasonably well.
\end{itemize}

\subsection{Built-in algorithms}


OPOF contains \emph{built-in} stable implementations of algorithms that solve the planner optimization problem, integrated directly into the core \code{opof} package. We expect this list to grow with time. 



\textbf{GC} is our implementation of the \emph{Generator-Critic} algorithm introduced across the recent works of \cite{lee2021a,lee2022a,danesh2022a}. Two neural networks are learned simultaneously -- a \emph{generator network} representing the generator \(G_\theta(c)\) and a \emph{critic network}. The generator is stochastic, and maps a problem instance \(c \in \mathcal{C}\) to a sample of planning parameters \(x \in \mathcal{X}\). The critic maps \(c\) and \(x\) into a real distribution modeling \(\boldsymbol{f}(x; c)\). During training, the stochastic generator takes a random problem instance \(c \in \mathcal{C}\) and produces a sample \(x \in \mathcal{X}\), which is used to probe the planner. The planner's response, along with \(c\) and \(x\), are used to update the critic via supervision loss. The critic then acts as a \emph{differentiable surrogate objective} for \(\boldsymbol{f}(x; c)\), passing gradients to the generator via the chain rule.

\textbf{SMAC} is a wrapper around the SMAC3 package \cite{lindauer2022smac3}, an actively maintained tool for algorithm configuration using the latest Bayesian optimization techniques. We use the \code{HPOFacade} strategy provided by SMAC3. In the context of the planner optimization problem, SMAC learns only a generator \(G_\theta(c) = \theta\) that is unconditional (i.e. does not change with the problem instance) and deterministic (i.e., always returns the same planning parameters). While SMAC does not exploit information specific to each problem instance, it provides an approach that has strong theoretical grounding and serves as a reasonable baseline and sanity check.

\section{Design Decisions}

\subsection{Many domains and many algorithms}
OPOF is domain-centric -- we only impose what a domain should \emph{look} like, not \emph{how} it should be solved. Such a choice maximizes the convenience of developing both domains and algorithms separately, by implicitly enforcing components to be reusable. One can easily reuse existing algorithms on a new domain of interest; similarly, one can also reuse existing domains to benchmark a new algorithm of interest.


\subsection{Portable and convenient packaging}

OPOF makes extensive use of \code{Pip}, the Python package manager, to allow external domains and algorithms to be packaged and distributed in a portable and convenient format. As such, users may, for example, easily install OPOF along with all available domain packages (at the time of OPOF's beta release) with: \codeg{\$ pip install opof opof-grid2d opof-sbmp opof-pomdp}.

\section{Future Directions}

We hope to apply/extend the POP formulation and the OPOF software framework in several ways.

\begin{itemize}
  \item \textbf{Interesting domain applications.} POP is a very general formulation that can be applied to a vast range of planners. With OPOF, we hope that this will ease such efforts for planning researchers interested in optimizing planners.
  \item \textbf{Developing general algorithms.} Right now, there is only one practical algorithm, the Generator-Critic (GC) algorithm, which is designed specifically for planner optimization. We hope that OPOF will aid research efforts to produce better and more efficient algorithms.
  \item \textbf{Harder parameter classes.} Existing domains have parameter spaces whose dimensions are reasonably small (\(<1000\)). We would like to explore how this framework scales to larger parameter spaces, and the domains for which this would be useful.
  \item \textbf{Applications outside of optimizing speed.} The definition of the planning objective is general, and allows us to apply POP outside of improving planning speed. For example, can we use it to maximize some form of safety or exploration? Or, in an entirely different field such as HRI, can we instead seek to find internal parameters maximizing the observed behavior of some non-differentiable agent?  
\end{itemize}

\section*{Acknowledgment}
This research is supported in part by NSF RI 2008720 and Rice University funds; by Shanghai Jiao Tong University Startup Funding No. WH220403042; and by the National Research Foundation (NRF), Singapore and DSO National Laboratories under the AI Singapore Program (AISG Award No: AISG2-RP-2020-016).

\printbibliography{}

\newpage 

\begin{appendices}

\section{Detailed Domain Descriptions}

\subsection{Grid world}\label{desc:grid2d}

\begin{figure}[h]
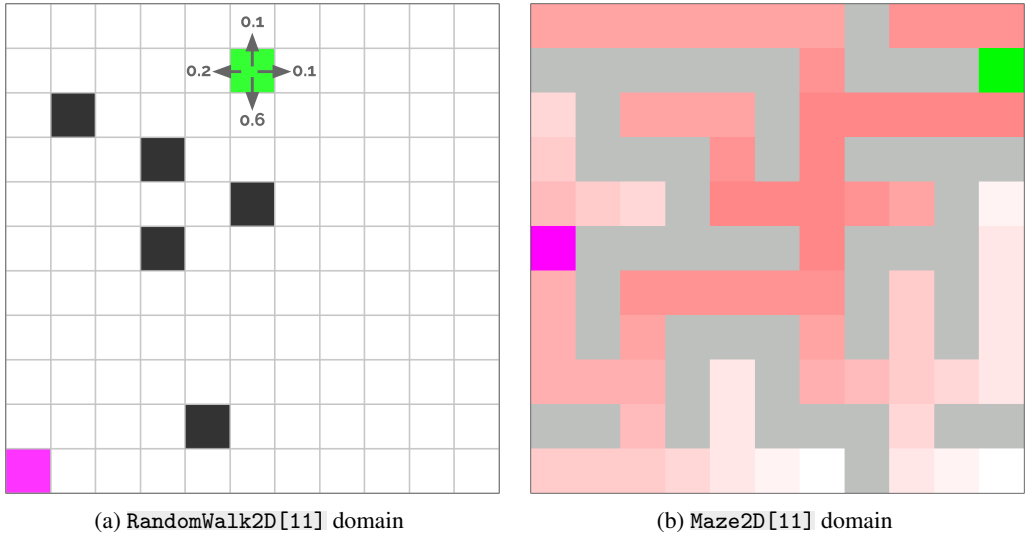

  \centering
  \begin{tabularx}{\linewidth}{YY}
    \begin{subfigure}[b]{1\linewidth}
      \includesvg[width=\linewidth]{figures/random_walk2d.svg}
      \caption{
        \code{RandomWalk2D[11]} domain
        \label{fig:randomwalk2d}
      }
    \end{subfigure}&
    \begin{subfigure}[b]{1\linewidth}
      \includesvg[width=\linewidth]{figures/maze2d.svg}
      \caption{
        \code{Maze2D[11]} domain
        \label{fig:maze2d}
      }
    \end{subfigure}
  \end{tabularx}
  \caption{Grid world domains.}
  \label{fig:gridworld}
\end{figure}

\textbf{Overview.} Through the \code{opof-grid2d} package, we provide two relatively simple grid world domains (\cref{fig:gridworld}): \code{RandomWalk2D[\emph{size}]} and \code{Maze2D[\emph{size}]}. We emphasize that we want to discover high-quality planning parameters \emph{solely} by interacting with the planner, where we treat the planner as a closed black-box function.  Specifically, we do not attempt to handcraft parameters specialized to the planner's operation. The goal is to be able to develop general algorithms for the broader planner optimization problem, using the grid world domains \emph{merely} to test these general algorithms. These general algorithms would apply to harder domains for which the complexity of the planner prevents us from handcrafting specialized parameters. 

\textbf{Random walk.} In the \code{RandomWalk2D[\emph{size}]} domain (\cref{fig:randomwalk2d}), an agent starts at a location (green) and moves in random directions according to some fixed probabilities until it reaches the goal (magenta). When attempting to move ``into'' an obstacle (black) or the borders of the grid, a step is spent but the position of the agent does not change. The probability of moving in each direction is fixed across all steps. \textbf{The planner optimization problem} is to find a generator \(G_\theta(c)\) that maps a problem instance \(c\) (in this case, the combination of board layout and start and goal positions) into direction probabilities (in this case, vectors \(\in \mathbb{R}^4\) with non-negative entries summing to \(1\)), such that the number of steps taken during the random walk is minimized. The training set and testing set each contain \(1000\) problem instances, where the obstacle, start, and goal positions are uniformly sampled.

\textbf{Maze A* search.} In the \code{Maze2D[\emph{size}]} domain (\cref{fig:maze2d}), A* search \cite{hart1968astar} is run against a \emph{heuristic} function \(h(n)\) to find a path from the start (green) to the goal (green). The heuristic function determines the priority in which nodes are expanded (cells that are in darker red have a lower \(g(n) + h(n)\) value, and have higher priority). The maze is assumed to be \emph{perfect}, i.e., there is exactly one path between any two cells. \textbf{The planner optimization problem} is to find a generator \(G_\theta(c)\) that maps a problem instance \(c\) (in this case, the combination of board layout and start and goal positions) to \(h(n)\) (in this case, assignments of values \(\in [0, size^2]\) to each of the \(size^2\) cells) such that the number of nodes expanded in the A* search is minimized. The training set and testing set each contain \(1000\) problem instances, where the maze is generated using Wilson's algorithm \cite{wilson1996maze}, and the start and goal positions are uniformly sampled.

\newpage

\subsection{Sampling-based motion planning (SBMP)}\label{desc:sbmp}

\begin{figure}[h!]
  \centering
  \begin{tabularx}{\linewidth}{YYY}
    \begin{subfigure}[b]{0.95\linewidth}
      \includegraphics[width=\linewidth]{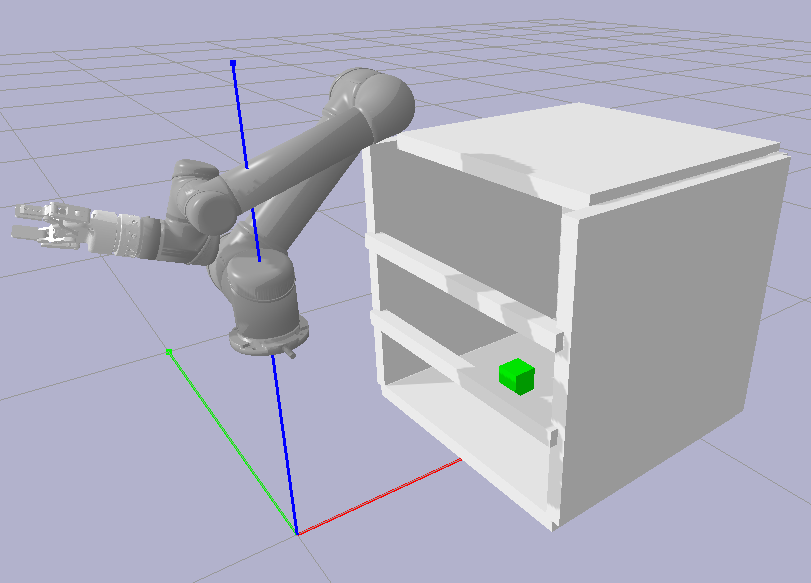}
      \caption{
        \code{Cage} (start)
        \label{fig:cagestart}
      }
    \end{subfigure}&
    \begin{subfigure}[b]{0.95\linewidth}
      \includegraphics[width=\linewidth]{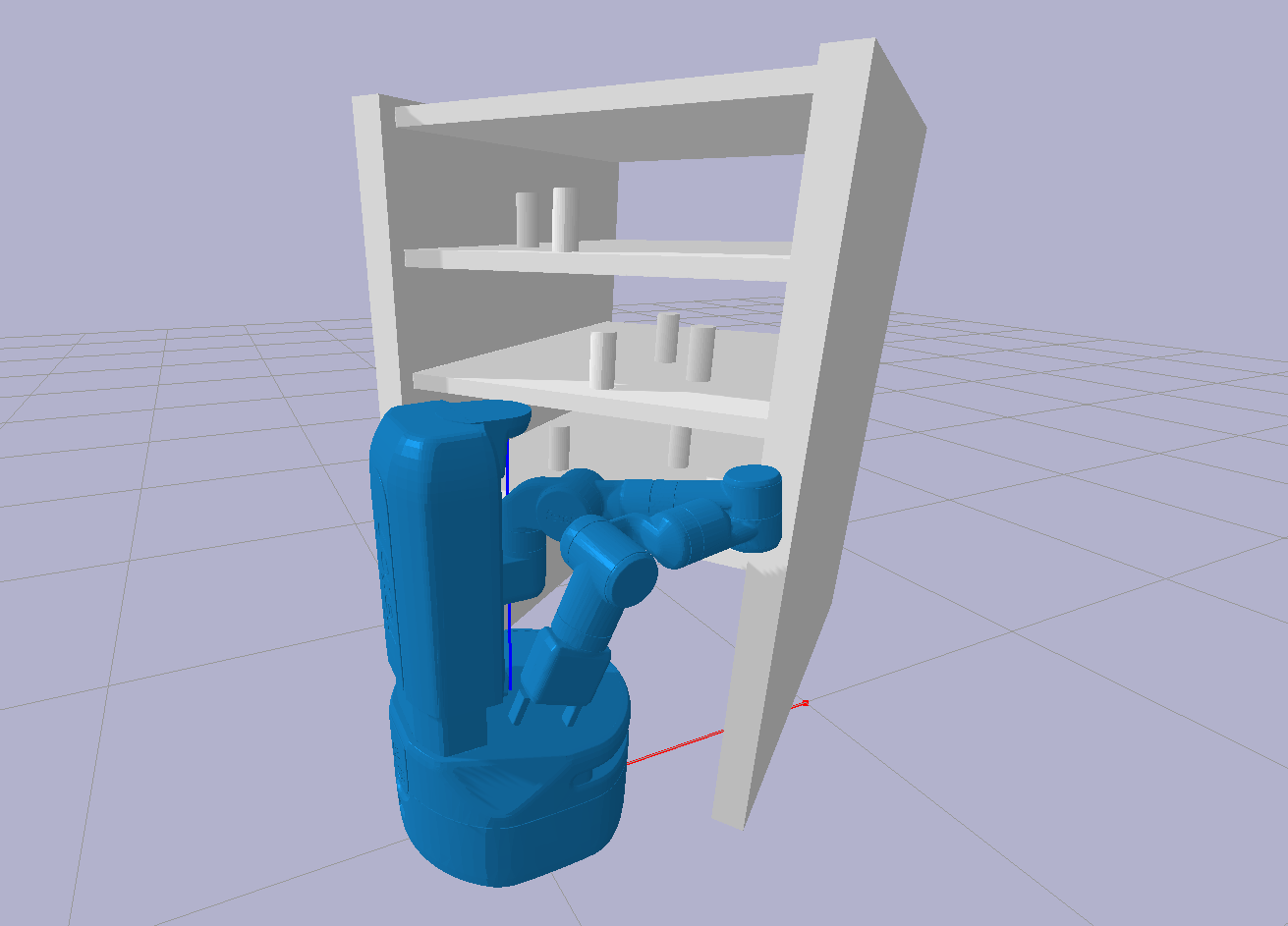}
      \caption{
        \code{Bookshelf} (start)
        \label{fig:bookshelfstart}
      }
    \end{subfigure}&
    \begin{subfigure}[b]{0.95\linewidth}
      \includegraphics[width=\linewidth]{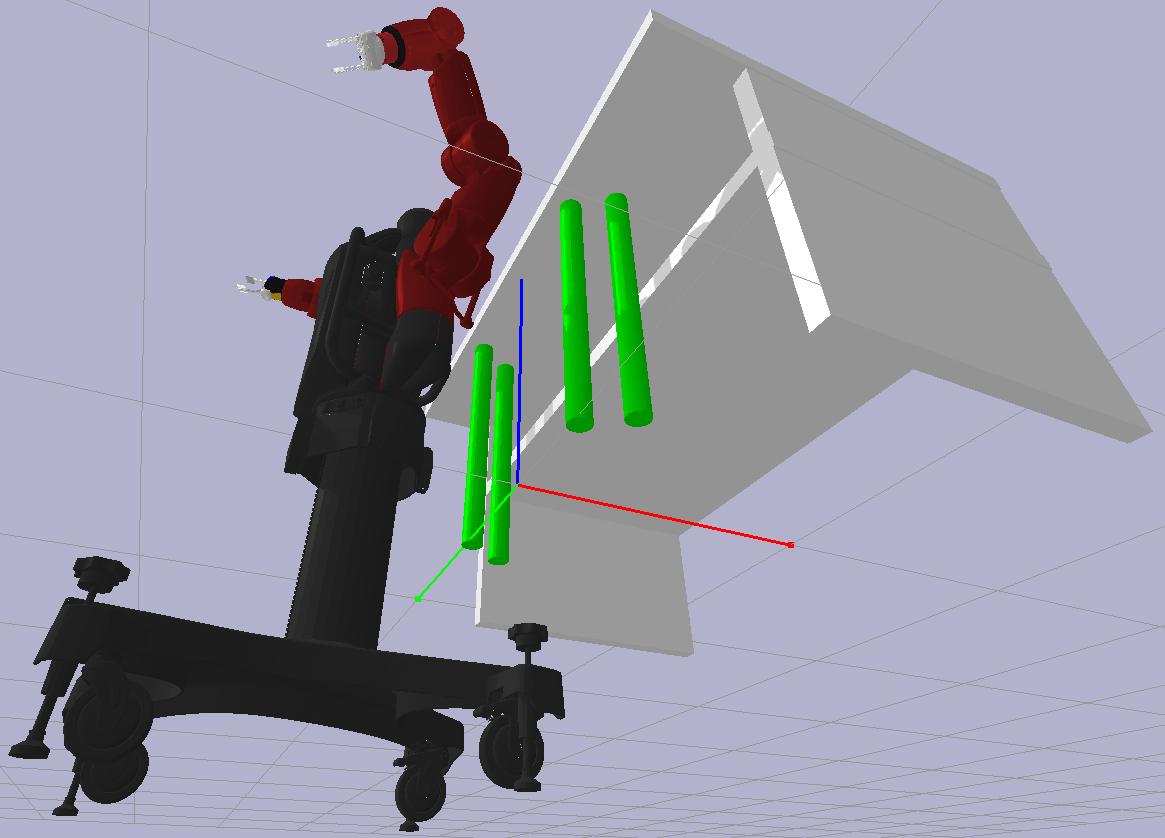}
      \caption{
        \code{Table} (start)
        \label{fig:tablestart}
      }
    \end{subfigure}
    \\\\

    \begin{subfigure}[b]{0.95\linewidth}
      \includegraphics[width=\linewidth]{figures/cage_goal.png}
      \caption{
        \code{Cage} (goal)
        \label{fig:cagegoal}
      }
    \end{subfigure}&
    \begin{subfigure}[b]{0.95\linewidth}
      \includegraphics[width=\linewidth]{figures/bookshelf_goal.png}
      \caption{
        \code{Bookshelf} (goal)
        \label{fig:bookshelfgoal}
      }
    \end{subfigure}&
    \begin{subfigure}[b]{0.95\linewidth}
      \includegraphics[width=\linewidth]{figures/table_goal.png}
      \caption{
        \code{Table} (goal)
        \label{fig:tablegoal}
      }
    \end{subfigure}\\
  \end{tabularx}
  \caption{Environments available in the \code{SBMPHyp[\emph{env},\emph{planner}]} domain. The start (top row) and goal (bottom row) robot configurations correspond to environment-specific tasks.}
  \label{fig:sbmp}
\end{figure}

\textbf{Overview.} Through the \code{opof-sbmp} package, we provide the \code{SBMPHyp[\emph{env},\emph{planner}]} domain (\cref{fig:sbmp}), which explores doing sampling-based motion planning in a specified environment using a specified planner. Unlike the grid world domains, these planners are much more complex and the relationship between the choice of planning parameters and planner performance is unclear. This makes it particularly suitable for OPOF, since we treat the planner as closed black-box function and specifically assume no knowledge of the planner's internals.

\textbf{The planner optimization problem.} The robot is tasked with moving its arm(s) from a start configuration to a goal configuration by running a sampling-based motion planner. The planner optimization problem is to find a generator \(G_\theta(c)\) that maps a problem instance \(c\) (in this case, the combination of obstacle poses in the environment and the start and goal robot configurations) to a set of planner hyperparameters (which depend on the planner used), such that the time taken for the motion planner to find a path is minimized. The training set and testing set contain \(1000\) and \(100\) problem instances respectively. These problem instances are adapted from MotionBenchMaker \cite{chamzas2021motionbenchmaker}. Obstacle positions are sampled according to a predefined distribution, while start and goal configurations are sampled using inverse kinematics \cite{buss2004ik} for some environment-specific task.

\textbf{Environments.} We provide \(3\) environments at the time of writing. 
In \code{Cage}, a \(6\)-dof UR5 robot is tasked to pick up a block (green) in a cage (\cref{fig:cagegoal}), starting from a random robot configuration (\cref{fig:cagestart}). The position and orientation of the cage, as well as the position of the block in the cage, are randomized. 
In \code{Bookshelf}, a \(8\)-dof Fetch robot is tasked to reach for a cylinder in a bookshelf (\cref{fig:bookshelfgoal}), starting from a random robot configuration (\cref{fig:bookshelfstart}). The position and orientation of the bookshelf and cylinders, as well as the choice of cylinder to reach for, are randomized. 
In \code{Table}, a \(14\)-dof dual-arm Baxter robot must fold its arms crossed in a constricted space underneath a table and in between two vertical bars (\cref{fig:tablegoal}), starting from a random robot configuration (\cref{fig:tablestart}). The lateral positions of the vertical bars are randomized.

\textbf{Planners.} We provide \(2\) planners at the time of writing. 
For \code{RRTConnect} \cite{kuffner2000rrt}, we grow two random search trees from the start and the goal configurations toward randomly sampled target points in the free space, until the two trees connect. We tune the following parameters: a \emph{range} \(\in [0.01, 5.00]\) parameter, which limits how much to extend the trees at each step; and a \emph{weight} vector \(\in \mathbb{R}^{50}\) with non-negative entries summing to \(1\), which controls the sampling of target points using the experience-based sampling scheme adapted from \cite{lee2022a}. For \code{LBKPIECE1} \cite{sucan2009bkpiece,bohlin2000lazyprm}, two random search trees are similarly grown from the start and the goal configurations, but controls the exploration of the configuration space using grid-based projections. We tune the following parameters: \emph{range}  \(\in [0.01, 5.00]\), \emph{border\_fraction} \(\in [0.001, 1]\), and \emph{min\_valid\_path\_fraction} \(\in [0.001, 1]\), which respectively determine how much to extend the trees at each step, how much to focus exploration on unexplored cells, and the threshold for which partially valid extensions are allowed; and a \emph{projection} vector \([0, 1]^{2 \times d} \subset \mathbb{R}^{2 \times d}\) which corresponds to the linear projection function used to induce the 2-dimensional exploration grid, where \(d\) is the robot's number of degrees of freedom.

\subsection{Online POMDP planning}\label{desc:pomdp}

\begin{figure}[h]
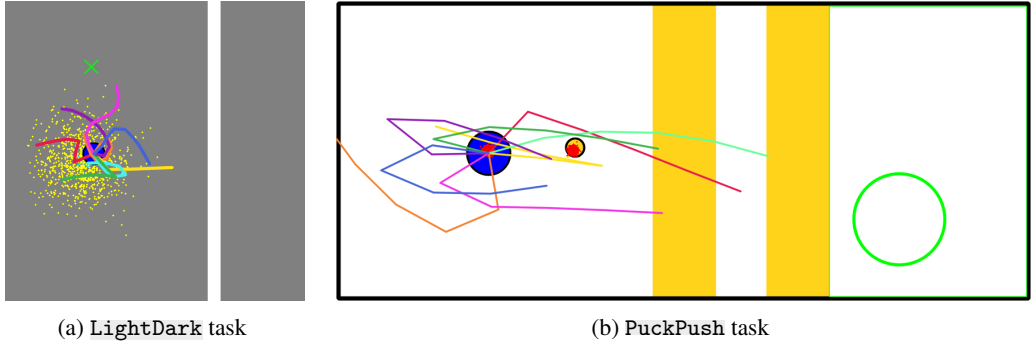

  \centering
  \begin{tabularx}{\linewidth}{cc}
    \begin{subfigure}[b]{0.285\linewidth}
      \includegraphics[width=\textwidth]{figures/lightdark_start.png}
      \caption{
        \code{LightDark} task
        \label{fig:lightdark}
      }
    \end{subfigure}&
    \begin{subfigure}[b]{0.66\linewidth}
      \includegraphics[width=\textwidth]{figures/puckpush_start.png}
      \caption{
        \code{PuckPush} task
        \label{fig:puckpush}
      }
    \end{subfigure}\\
  \end{tabularx}
  \caption{Planning tasks available in the \code{POMDPMacro[\emph{task},\emph{length}]} domain.}
  \label{fig:pomdp}
\end{figure}

\textbf{Overview.} Through the \code{opof-pomdp} package, we provide the \code{POMDPMacro[\emph{task},\emph{length}]} domain, which explores doing online POMDP planning for a specified task using the DESPOT \cite{ye2017despot} online POMDP planner. The robot operates in a \emph{partially observable} world, and tracks a belief over the world's state across actions that it has taken. Given the current belief at each step, the robot must determine a good action (which corresponds to moving a fixed distance toward some heading) to execute. 
It does so by running the DESPOT online POMDP planner. DESPOT runs some form of anytime Monte-Carlo tree search over possible action and observation sequences, rooted at the current belief, and returns a lower bound for the computed partial policy. 

\textbf{The planner optimization problem.} Since the tree search is exponential in search depth, \code{POMDPMacro[\emph{task},\emph{length}]} explores using open-loop \emph{macro-actions} to improve the planning efficiency. Here, DESPOT is parameterized with a set of \(8\) \emph{macro-actions}, which are 2D cubic Bezier curves stretched and discretized into \(length\) number of line segments that determine the heading of each corresponding action in the macro-action. Each Bezier curve is controlled by a \emph{control} vector \(\in \mathbb{R}^{2 \times 3}\), which determine the control points of the curve. Since the shape of a Bezier curve is invariant up to a fixed constant across the control points, we constrain the control vector to lie on the unit sphere. The planner optimization problem is to find a generator \(G_\theta(c)\) that maps a problem instance (in this case, the combination of the current belief, represented as a particle filter, and the current task parameters, whose representation depends on the task) to a \emph{joint control} vector \(\in \mathbb{R}^{8 \times 2 \times 3}\) (which determines the shape of the \(8\) macro-actions), such that the lower bound value reported by DESPOT is maximized. \code{POMDPMacro[\emph{task},\emph{length}]} comes with \(2\) tasks at the time of writing.

\textbf{Light-Dark.} In the \code{LightDark} task (\cref{fig:lightdark}), the robot (blue circle) wants to move to and stop exactly at a goal location (green cross). However, it cannot observe its own position in the dark region (gray background), but can do so only in the light region (white vertical strip). It starts with uncertainty over its position (yellow particles) and should discover, through planning, that localizing against the light before attempting to stop at the goal will lead to a higher success rate, despite taking a longer route. The task is parameterized by the goal position and the position of the light strip, which are uniformly selected.

\textbf{Puck-Push.} In the \code{PuckPush} task (\cref{fig:puckpush}), a circular robot (blue) pushes a circular puck (yellow circle) toward a goal (green circle). The world has two vertical strips (yellow) which have the same color as the puck, preventing observations of the puck from being made when on top. The robot starts with little uncertainty (red particles) over its position and the puck's position corresponding to sensor noise, which grows as the puck moves across the vertical strips. Furthermore, since both robot and puck are circular, the puck slides across the surface of the robot whenever it is pushed. The robot must discover, through planning, an extremely long-horizon plan that can (i) recover localization of the puck, and can (ii) recover from the sliding effect by retracing to re-push the puck. The task is parameterized by position of the goal region, which is uniformly selected within the white area on the right.

\code{POMDPMacro[\emph{task},\emph{length}]} differs from the other domain in that the distribution of problem instances \(\mathcal{D}\) is \emph{dynamic}. It is hard to prescribe a ``dataset of beliefs'' in online POMDP planning to construct a problem instance distribution. The space of reachable beliefs is too hard to determine beforehand, and too small relative to the entire belief space to sample at random. Instead, \code{POMDPMacro[\emph{task},\emph{length}]} loops through episodes of planning and execution, returning the current task parameters and belief at the current step whenever samples from \(\mathcal{D}\) are requested. Such a dynamic distribution adds noise to the training procedure, causing it to slow down. We believe that in the future, accounting for this noise will likely speed up training.

\section{Experimental Configuration}

We run GC and SMAC against across ranges of domains. All experiments were done on a machine with an AMD 5900x (24 threads @ 3.70 GHz), an RTX 3060, and 32GB of RAM. The choice of planning objective and range of training domain configurations are detailed as follows:
\begin{itemize}
  \item {\code{RandomWalk2D[\emph{size}]}}: The planning objective \(\boldsymbol{f}(x; c)\) is given as \(- steps / (4 \times size^2)\), where \(steps\) is the number of steps taken to reach the goal. A maximum of \(4 \times size^2\) steps are allowed. We vary \(\mathrm{\code{\emph{size}}} = 5, 11, 21\) and allow up to \(200\)K planner calls for training. We use \(8\) worker threads for training.
  
  \item {\code{Maze2D[\emph{size}]}}: The planning objective \(\boldsymbol{f}(x; c)\) is given as \(- steps / n_{\mathrm{empty}}\), where \(steps\) is the number of nodes expanded before finding the goal and \(n_{\mathrm{empty}}\) is the number of obstacle-free cells. We vary \(\mathrm{\code{\emph{size}}} = 5, 11, 21\) and allow up to \(150\)K planner calls for training.  We use \(8\) worker threads for training.
  
  \item {\code{SBMPHyp[\emph{env},\emph{planner}]}}: The planning objective \(\boldsymbol{f}(x; c)\) is given as \(-time\), where \(time\) is the time taken for the motion planner to find a collision-free path from the start to goal robot configuration. We vary \(\mathrm{\code{\textit{env}}} = \mathrm{\code{Cage}}, \mathrm{\code{Bookshelf}}, \mathrm{\code{Table}}\) with maximum allowed planning times of \(1, 2, 3\) seconds respectively, and vary \(\mathrm{\code{\textit{planner}}} = \mathrm{\code{RRTConnect}}, \mathrm{\code{LBKPIECE1}}\). We allow up to \(100\)K planner calls for training, and use \(16\) worker threads.
  
  \item {\code{POMDPMacro[\emph{env},\emph{planner}]}}: The planning objective \(\boldsymbol{f}(x; c)\) is given as the lower bound value reported by DESPOT, under a timeout of \(100\) ms. When evaluating, we instead run the planner across \(50\) episodes, at each step calling the generator, and compute the average sum of rewards (as opposed to considering the lower bound value for a single belief during training). We vary \(\mathrm{\code{\textit{env}}} = \mathrm{\code{LightDark}}, \mathrm{\code{PuckPush}}\), using \(\mathrm{\code{\textit{length}}} = 8, 5\) respectively, which are suggested to be the optimal macro-action lengths in \cite{lee2021a}. We allow up to \(500\)K and \(1\)M planner calls respectively for training, and use \(16\) worker threads.
\end{itemize}

When running GC, we update the generator for every sample of problem instance, planning parameters, and planner call. For SMAC, which does not condition the generator on the problem instance, we instead sample a set of planning parameters, test its average performance across \(50\) randomly sampled problem instances, and use the average result reported by the planner to update the generator. This is because SMAC's runtime grows cubically in the number of updates --- this strategy allows us to use the same number of planner calls but limit the number of actual updates. We instead attempt to make each update more efficient by averaging the results of the planner calls to reduce noise.

\begin{table}[t]
  \caption{Key domain properties and planning performance after training. \(\dim(\mathcal{C})\) refers to the dimensionality of the space of problem instances. \(\dim(\mathcal{X})\) refers to the dimensionality of the space of planning parameters. \(\mathrm{Range}(\mathrm{Perf.})\) refers to the achievable range of the planning performance. \(\mathrm{Perf.}\) refers to the evaluated planning performance after training (we average this over the last \(5\) generator training checkpoints). \(t\) refers to the training time in hours.}
  \label{table:planning}
  \vspace{8pt}
  \centerline{
    \renewcommand{\arraystretch}{1.1}
    \begin{tabular}{rcccccccc}
    \toprule\toprule
    \multicolumn{1}{c}{} & \multicolumn{3}{c}{Properties} & \multicolumn{2}{c}{GC} & \multicolumn{2}{c}{SMAC} \\

    \cmidrule(lr){2-4}\cmidrule(lr){5-6}\cmidrule(lr){7-8}

    Domain & {\footnotesize\(\dim(\mathcal{C})\)} & {\footnotesize\(\dim(\mathcal{X})\)} & {\footnotesize\(\mathrm{Range}(\mathrm{Perf.})\)} & {\footnotesize \(\mathrm{Perf.}\)} & {\footnotesize \(t\)} & {\footnotesize \(\mathrm{Perf.}\)} & {\(t\)} \\

    \midrule

    \codes{RandomWalk2D[5]} & 29 & 4 & \([-1, 0]\) & \textbf{-0.518} & 1.1 & -0.905 & 0.6 \\
    \codes{RandomWalk2D[11]} & 125 & 4 & \([-1, 0]\) & \textbf{-0.749} & 1.1 & -0.946 & 0.7 \\
    \codes{RandomWalk2D[21]} & 445 & 4 & \([-1, 0]\) & \textbf{-0.856} & 1.2 & -0.966 & 1.0 \\
    \midrule
    \codes{Maze2D[5]} & 29 & 25 & \([-1, 0]\) & \textbf{-0.379} & 0.5 & -0.524 & 1.6 \\
    \codes{Maze2D[11]} & 125 & 121 & \([-1, 0]\) & \textbf{-0.468} & 0.5 & -0.545 & 8.3 \\
    \codes{Maze2D[21]} & 445 & 441 & \([-1, 0]\) & \textbf{-0.472} & 0.5 & -0.508 & 12.3\textsuperscript{\hyperref[footnote:earlyterm1]{1}} \\
    \midrule
    \codes{SBMPHyp[Cage,RRTConnect]} & 26 & 51 & \([-1, 0]\) & \textbf{-0.473} & 1.1 & -0.604 & 3.0 \\
    \codes{SBMPHyp[Cage,LBKPIECE1]} & 26 & 15 & \([-1, 0]\) & -0.695 & 1.5 & \textbf{-0.619} & 2.3 \\
    \codes{SBMPHyp[Bookshelf,RRTConnect]} & 86 & 51 & \([-2, 0]\) & \textbf{-1.21} & 2.4 & -1.31 & 4.9 \\
    \codes{SBMPHyp[Bookshelf,LBKPIECE1]} & 86 & 19 & \([-2, 0]\) & -1.09 & 2.2 & \textbf{-1.04} & 3.6 \\
    \codes{SBMPHyp[Table,RRTConnect]} & 30 & 51 & \([-3, 0]\) & \textbf{-1.24} & 2.7 & -2.05 & 6.4 \\
    \codes{SBMPHyp[Table,LBKPIECE1]} & 30 & 31 & \([-3, 0]\) & \textbf{-2.88} & 6.1 & -2.92 & 8.1 \\
    \midrule
    \codes{POMDPMacro[LightDark,8]} & 50,002 & 48 & \([-106, 100]\) & \textbf{39.4} & 4.4 & 11.0 & 9.3 \\
    \codes{POMDPMacro[PuckPush,5]} & 30,003 & 48 & \([-110, 100]\) & \textbf{78.6} & 17.0 & 66.8 & 25.8\textsuperscript{\hyperref[footnote:earlyterm2]{2}}\\

    \bottomrule\\
  \end{tabular}
  }
\end{table}

\section{Experimental Results}

Training results are shown in \cref{table:training}. After training, we evaluate the trained generators against the training problem distribution, and show the results in \cref{table:planning}. We also show the key properties of each domain, including the dimensionalities of the space of problem instance and the space of planning parameters. 

\newpage 
\textbf{Conditioning improves performance.} We note from \cref{table:planning} and \cref{table:training} that in general, GC converges with better performance than SMAC. This is expected, since GC learns a conditional generator which can produce better parameters by exploiting information about the specific problem instance. On the other hand, SMAC maintains the AC formulation of optimization by looking for a single set of parameters that works well on average across the entire training set. It generally performs poorly, since it does not exploit information specific to each problem instance. In rare cases, SMAC is able to perform as well as GC (e.g. \code{SBMPHyp[Bookshelf,LBKPIECE1]} and \code{POMDPMacro[PuckPush,5]}), indicating that the additional information of individual problem instances do no provide any benefit to GC. SMAC occasionally outperforms GC (e.g. \code{SBMPHyp[Cage,LBKPIECE1]}), suggesting that constraining the solution space to be simple (i.e., to the set of unconditional generators) may make it easier to find a good solution.

\textbf{Not conditioning converges faster.} From \cref{table:training}, SMAC generally converges faster (albeit with poorer performance). This is most noticeable in cases such as \code{SBMPHyp[Table,RRTConnect]}, \code{SBMPHyp[Bookshelf,LBKPIECE1]}, and \code{POMDPMacro[PuckPush,5]}. This is again expected, since the space of generators considered by GC is strictly larger than that of SMAC, increasing the exploration time needed for GC before converging. 

\textbf{Fundamental optimization limits.} On the toy domains \code{RandomWalk[\emph{size}]} and \code{Maze2D[\emph{size}]}, GC almost completely fails to improve performance once the domain size becomes large enough. This hints that GC has inherent limits with respect to the structure of the problem. Does it fail, because of sample complexity issues, or because of structural priors over the parameter space, or something else? Since these toy domains are relatively well-behaved and easy to understand, there is much room for potential theoretical analysis.

\blfootnote{\textsuperscript{1, 2}\label{footnote:earlyterm1}\label{footnote:earlyterm2} Terminated early after 58K and 500K planner calls respectively to keep training times reasonable. Due to the cubic 
nature of SMAC, the full duration would have required approx. 2.6 and 8.6 days respectively.}

\begin{longtable}{lll}
    \caption{Training performance. We evaluate the generator at regular intervals across checkpoints, and plot the measured planning performance.}
    \label{table:training}\\

    \includegraphics[width=0.3\linewidth]{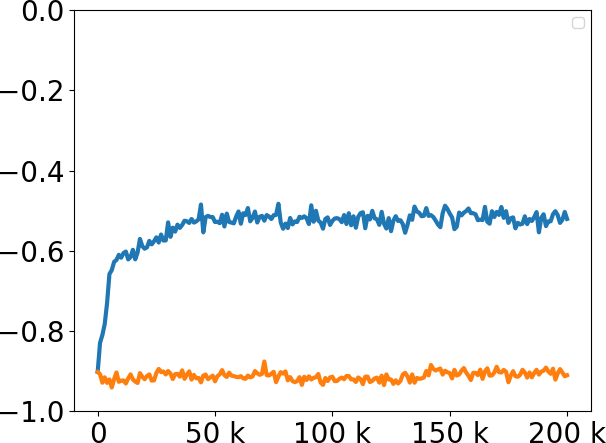}
    &
    \includegraphics[width=0.3\linewidth]{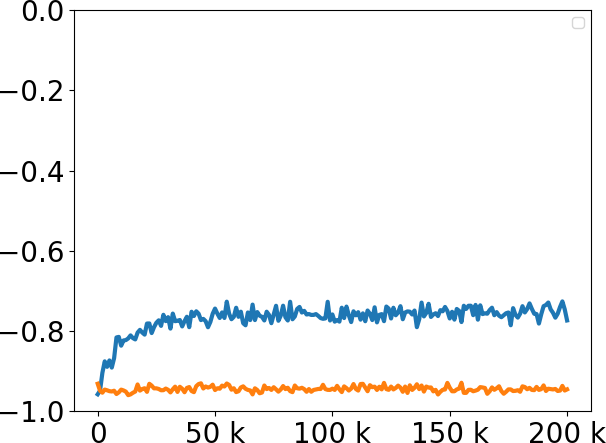}
    &
    \includegraphics[width=0.3\linewidth]{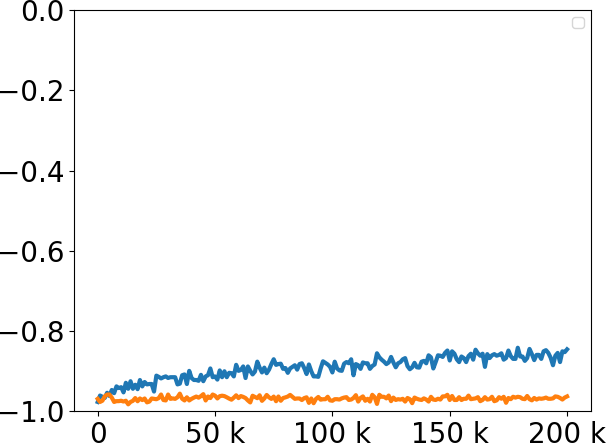}
    \\

    \multicolumn{1}{c}{\codess{RandomWalk2D[5]}}
    &
    \multicolumn{1}{c}{\codess{RandomWalk2D[11]}}    
    &
    \multicolumn{1}{c}{\codess{RandomWalk2D[21]}}
    \\\\

    \includegraphics[width=0.3\linewidth]{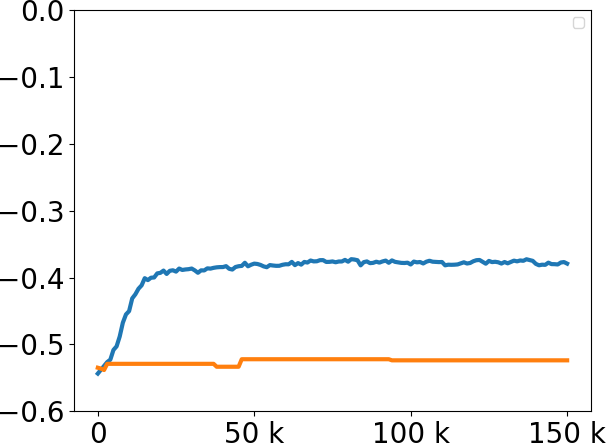}
    &
    \includegraphics[width=0.3\linewidth]{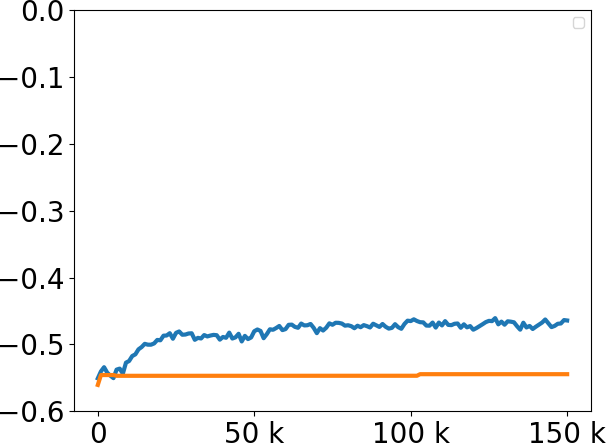}
    &
    \includegraphics[width=0.3\linewidth]{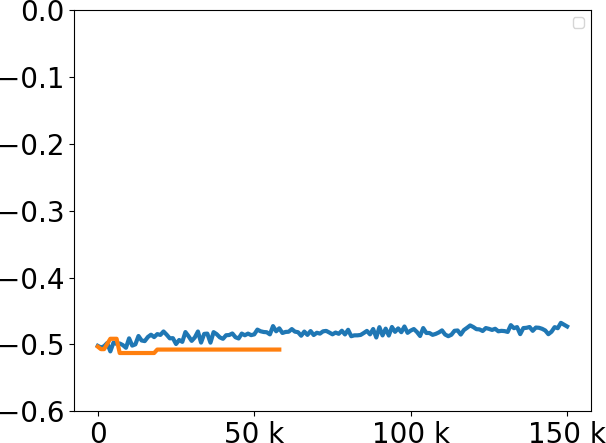}
    \\

    \multicolumn{1}{c}{\codess{Maze2D[5]}}
    &
    \multicolumn{1}{c}{\codess{Maze2D[11]}}
    &
    \multicolumn{1}{c}{\codess{Maze2D[21]}}
    \\\\
    
    \includegraphics[width=0.3\linewidth]{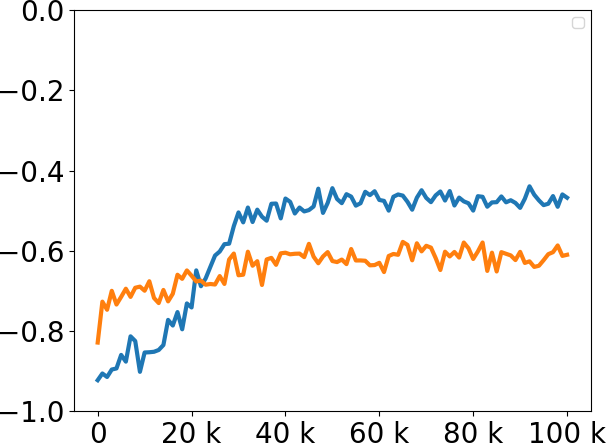}
    &
    \includegraphics[width=0.3\linewidth]{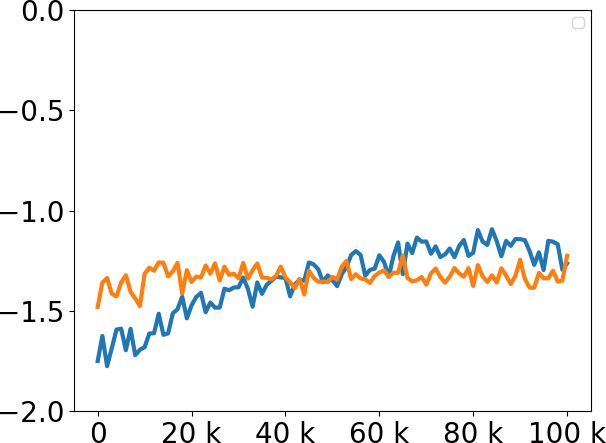}
    &
    \includegraphics[width=0.3\linewidth]{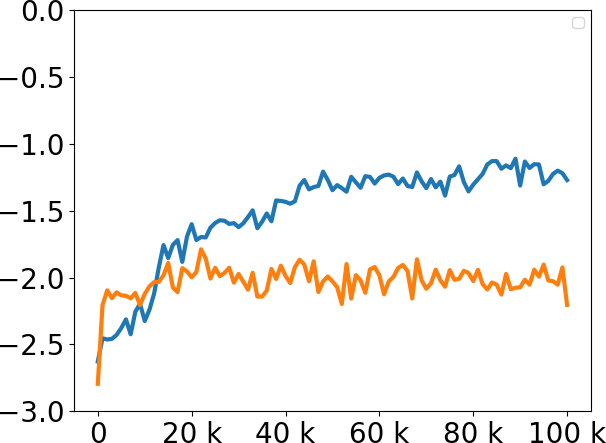}
    \\
    
    \multicolumn{1}{c}{\codess{SBMPHyp[Cage,RRTConnect]}}
    &
    \multicolumn{1}{c}{\codess{SBMPHyp[Bookshelf,RRTConnect]}}
    &
    \multicolumn{1}{c}{\codess{SBMPHyp[Table,RRTConnect]}}
    \\\\

    \includegraphics[width=0.3\linewidth]{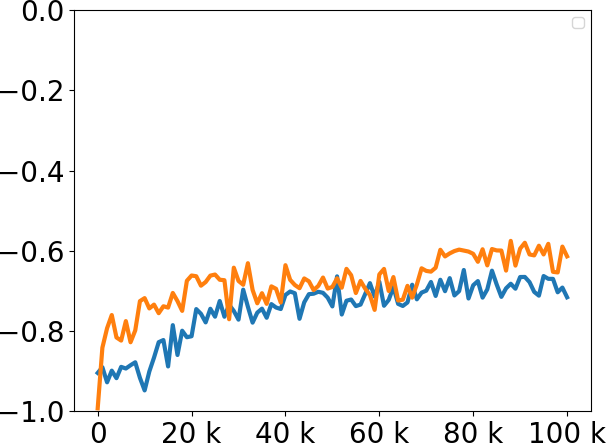}
    &
    \includegraphics[width=0.3\linewidth]{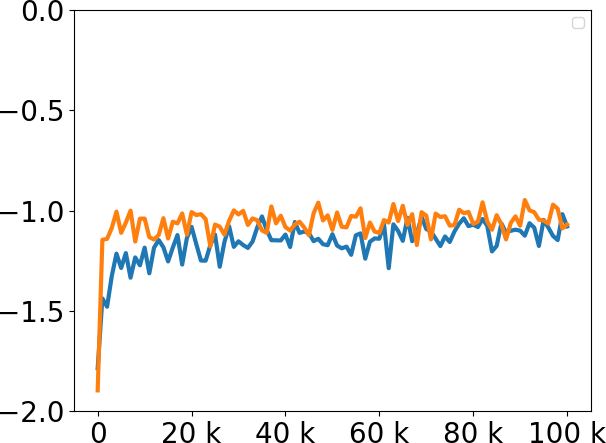}
    &
    \includegraphics[width=0.3\linewidth]{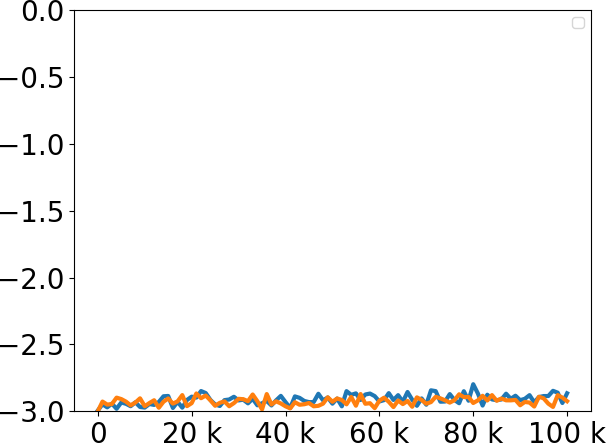}
    \\

    \multicolumn{1}{c}{\codess{SBMPHyp[Cage,LBKPIECE1]}}
    &
    \multicolumn{1}{c}{\codess{SBMPHyp[Bookshelf,LBKPIECE1]}}
    &
    \multicolumn{1}{c}{\codess{SBMPHyp[Table,LBKPIECE1]}}
    \\\\

    \includegraphics[width=0.3\linewidth]{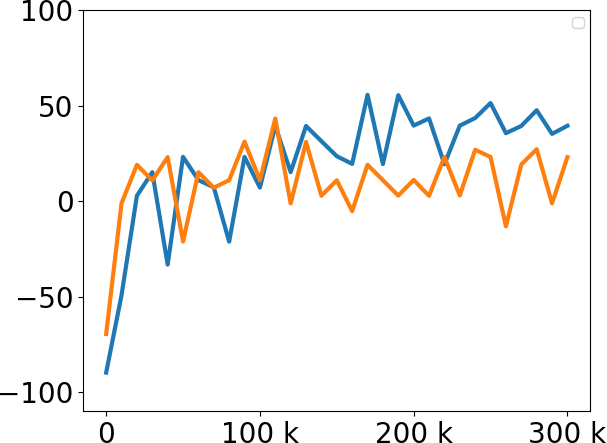}
    &
    \includegraphics[width=0.3\linewidth]{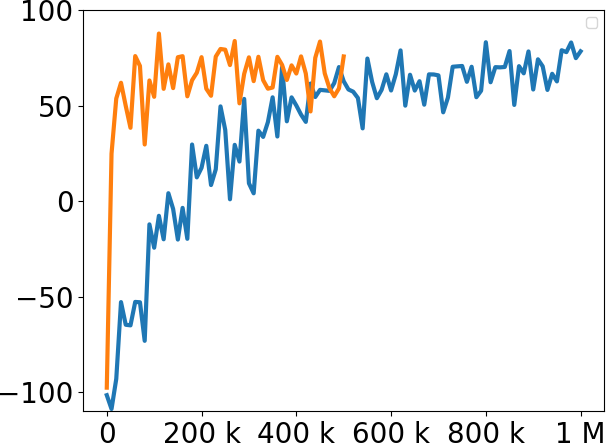}
    &
    \multicolumn{1}{c}{\multirow{1}{*}[55pt]{\includegraphics[width=0.18\linewidth]{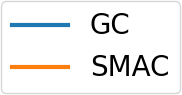}}}
    \\

    \multicolumn{1}{c}{\codess{POMDPMacro[LightDark,8]}}
    &
    \multicolumn{1}{c}{\codess{POMDPMacro[PuckPush,5]}}
    &

  \end{longtable}

\end{appendices}

\end{document}